\pdfoutput=1
\documentclass[10pt,twocolumn]{article}

\usepackage[letterpaper,margin=0.75in]{geometry}
\usepackage{amsmath,amssymb,amsfonts}
\usepackage{graphicx}
\usepackage{booktabs}
\usepackage{multirow}
\usepackage{cite}
\usepackage{hyperref}

\begin{document}


\title{DINO-GFSA: Geo-Localization via Semantic Gated Fusion and Mamba-based Sequential Aggregation}

\author{
    \begin{tabular}{c}
    Beier Hu, Yuanshen Guo, Jialu Cai, Chengwei Li,Yong Wang\textsuperscript{*}, Shunan Wu, and Zhigang Wu\\
    School of Aeronautics and Astronautics, Sun Yat-sen University, Shenzhen, China\\
    Email: \{huber, guoysh25, caijlu, lichw26\}@mail2.sysu.edu.cn\\
    \{wangyong5, wushunan, wuzhigang\}@mail.sysu.edu.cn
    \end{tabular}\\[0.6em]
    \parbox{0.92\textwidth}{\centering\small
    This work was jointly supported by the Basic Science Center Program of the National Natural Science Foundation of China (62388101) and the Innovation Fund for University-Industry-Research Collaboration of the Ministry of Education of China (Grant No. 2024ZY011).\\
    \textsuperscript{*}Corresponding author: Yong Wang.}
}
\date{}

\maketitle

\begin{abstract}
Cross-view geo-localization (CVGL) is critical for Unmanned Aerial Vehicle (UAV) self-positioning and target localization in GNSS-denied environments. However, acquiring robust semantics while preserving fine-grained spatial details remains challenging. To address this, we propose DINO-GFSA, a framework leveraging a LoRA (Low-Rank Adaptation) adapted DINOv3 (ViT-L) backbone for parameter-efficient, high-capacity representation. Crucially, we introduce a Semantic Gated Residual Fusion module, which utilizes high-level semantics to selectively calibrate and integrate low-level spatial cues, effectively bridging the semantic gap. Furthermore, a Mamba-based Sequential Aggregation Head is designed to capture long-range spatial dependencies with linear complexity. Experiments demonstrate state-of-the-art performance on University-1652 and DenseUAV benchmarks, notably surpassing the previous best on DenseUAV by 3.48\% on Recall@1. These results validate DINO-GFSA as a generalized, robust solution for UAV CVGL.
\end{abstract}


\section{Introduction}
\label{sec:intro}

Unmanned Aerial Vehicles (UAVs) have become indispensable in critical missions such as post-disaster reconstruction, owing to their remarkable flexibility and rapid response capabilities. However, these operations frequently necessitate deployment in Global Navigation Satellite System (GNSS)-denied environments, underscoring a critical need for robust, GNSS-independent navigation technologies. Concurrent with advancements in computer vision, visual-based geo-localization has emerged as a promising solution for UAV target localization \cite{Zheng2020University} and self-positioning \cite{Dai2023Vision, Wu2025UAV}. The core task involves pinpointing the UAV's location by matching real-time aerial imagery with a geo-referenced satellite database. Despite its potential, the significant discrepancies in viewpoint and visual appearance between UAV and satellite imagery remain a fundamental challenge.

While existing solutions have achieved promising accuracy, we identify critical limitations in how current methods handle feature extraction, fusion, and aggregation. \textbf{First}, regarding feature extraction, earlier methods primarily rely on standard backbones like ResNet \cite{Xia2024Enhancing} or Vision Transformers \cite{Dai2023Vision}. Although recent approaches utilizing Large Vision Models (LVMs) like DINOv2 \cite{Oquab2023Dinov2, Yang2025Dinov2} have shown promise, they still fail to exhibit a clear advantage over other backbones. \textbf{Second}, regarding feature fusion, deeper layers provide rich semantics while shallower layers retain spatial details. Existing methods often employ naive addition or use architectures like FPN \cite{Lin2017Feature}, which lack sufficient mechanisms to selectively combine semantic and spatial information. \textbf{Third}, regarding aggregation, simple pooling operations (e.g., Generalized Mean (GeM) Pooling \cite{Radenovic2018Fine}) often create an information bottleneck, discarding the complex spatial relationships essential for precise matching.

\begin{figure}[t] 
\centering
\includegraphics[width=\linewidth]{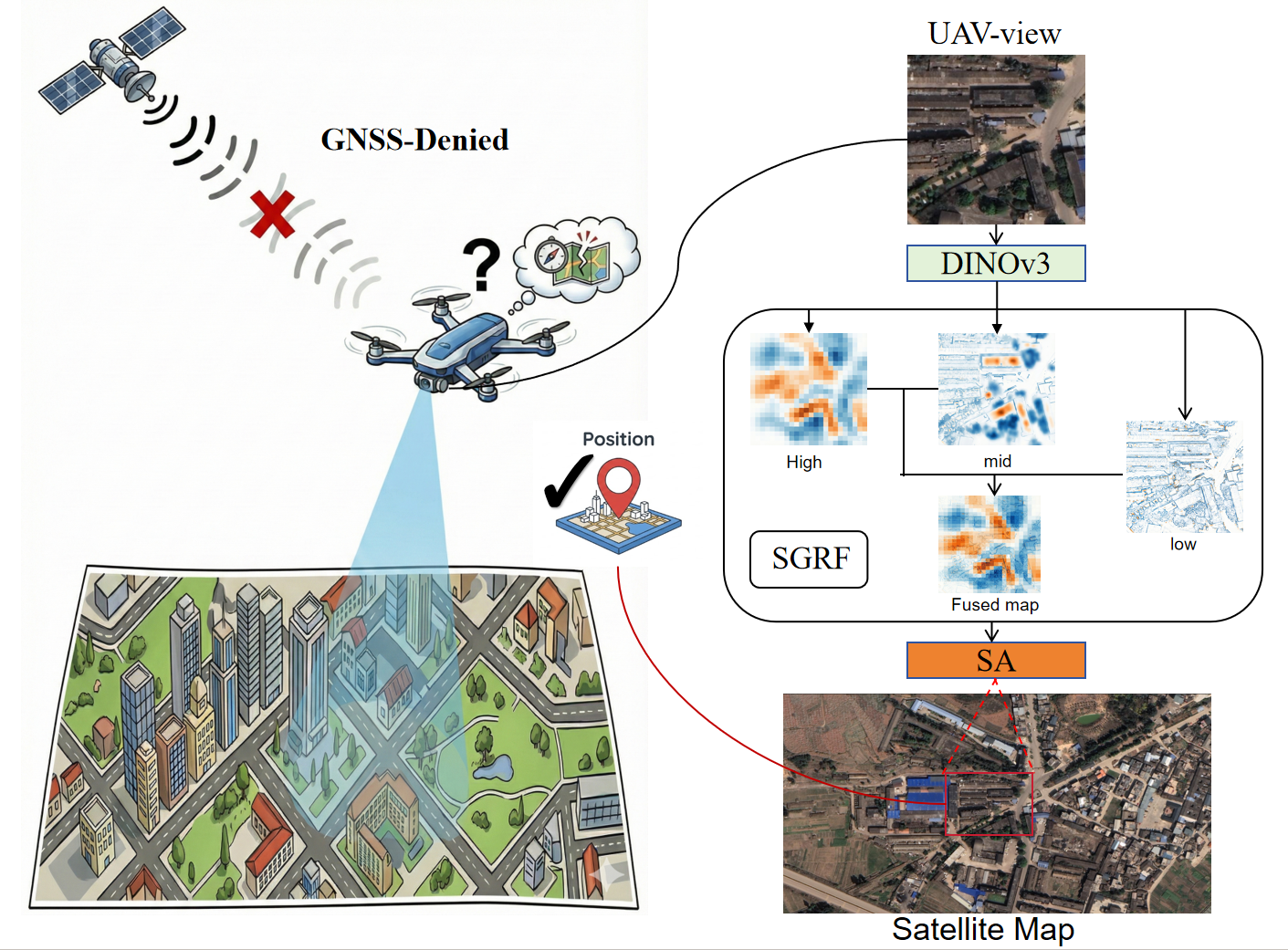} 
\caption{Concept and framework overview.}
\label{fig:concept}
\vspace{-0.6cm}
\end{figure}

To address these challenges, we propose DINO-GFSA, a comprehensive framework for cross-view geo-localization (CVGL), as illustrated in Fig.~\ref{fig:concept}.  The core of our approach consists of three key innovations: (1) \textbf{Backbone}: We constitute the first attempt to investigate DINOv3 \cite{Simeoni2025Dinov3} as the backbone for this task. To explore the \textit{performance upper bound} of LVMs, we adopt the {ViT-Large} architecture. To resolve the training difficulties of such a massive model, we utilize Low-Rank Adaptation (LoRA) \cite{Hu2022Lora}, achieving efficient adaptation with minimal trainable parameters ($\approx$ 45.73M) while preserving the robust generalization of the pre-trained weights. (2) \textbf{Fusion}: We propose the Semantic Gated Residual Fusion (SGRF) module, which uses high-level semantics to generate gating signals that \textit{selectively calibrate} low-level feature. This ensures that only task-relevant spatial details are integrated, effectively bridging the semantic gap. (3) \textbf{Aggregation}: We introduce a Sequential Aggregation (SA) Head based on Mamba \cite{Gu2024Mamba}. By treating the 2D feature map as a flattened sequence, it captures long-range dependencies with linear complexity.

The main contributions are summarized as follows:

\begin{itemize}
\item We are the first to investigate DINOv3 as a backbone for the CVGL task, utilizing a LoRA adaptation strategy to achieve high-capacity representation with minimal trainable parameters.
\item We design the SGRF module to bridge the semantic and spatial gap between hierarchical features, leveraging a semantic gating mechanism to selectively integrate fine-grained geometric cues.
\item We introduce a Mamba-based SA Head that treats feature maps as sequences to capture long-range spatial dependencies with linear complexity, effectively overcoming the limitations of traditional pooling.
\end{itemize}

\section{Methodology}

\subsection{Overall Architecture}
\begin{figure*}[t] 
    \centering
    \vspace{-0.2cm}
    \includegraphics[width=\linewidth]{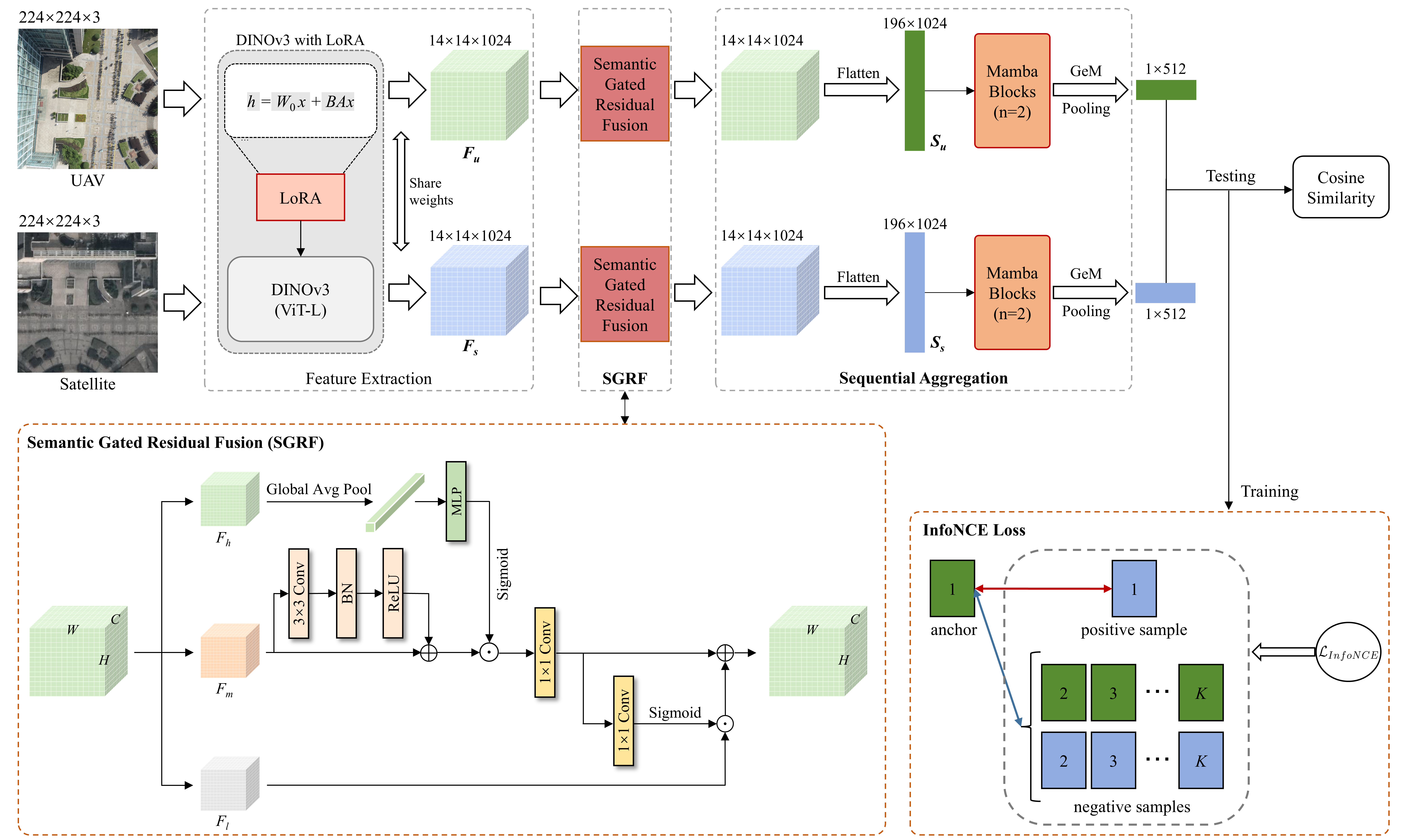}
    \caption{Overview of DINO-GFSA. The framework utilizes a LoRA-adapted DINOv3 backbone for multi-scale feature extraction. The SGRF module (bottom-left) is designed to filter low-level noise via semantic gating. Subsequently, the SA head captures global dependencies to generate the final descriptor. The network is trained end-to-end using InfoNCE loss (bottom-right).}
    \label{fig:arch}
    \vspace{-0.5cm}
\end{figure*}

The architecture of DINO-GFSA is presented in Fig.~\ref{fig:arch}. Adopting a Siamese network structure with shared weights, we employ DINOv3 (ViT-L) with LoRA as the backbone to extract robust features efficiently from both UAV and satellite views simultaneously. To fully exploit hierarchical information, feature maps from the 16th, 20th, and 24th layers are processed by the SGRF module, which effectively fuses distinct semantic and spatial properties. The fused representation is then flattened and passed to the SA Head, where stacked Mamba blocks model long-range spatial dependencies within the sequence. The framework concludes with GeM Pooling to generate the final feature vector. The model is trained in an end-to-end manner using Information Noise-Contrastive Estimation (InfoNCE) loss \cite{Oord2018Representation}, while cosine similarity is employed for matching during the testing phase.

\subsection{Backbone: DINOv3 with LoRA}
Feature extraction determines the upper limit of the model's matching performance. While Transformer-based backbones (e.g., ViT \cite{Dosovitskiy2020Image}, Swin \cite{Chen2025SHAA}) excel at capturing long-range dependencies, they are often hindered by the scarcity of domain-specific training data. To address this, we leverage Large Vision Models (LVMs), specifically adopting DINOv3 (ViT-L) \cite{Simeoni2025Dinov3} as our backbone. DINOv3 improves scalability and feature robustness over its predecessors through advanced regularization strategies. We intentionally select the ViT-Large variant (305M params) to explore the full performance potential of LVMs in processing complex cross-view imagery.

However, efficiently adapting such a massive model remains a challenge, as full fine-tuning is computationally prohibitive and risks catastrophic forgetting. To strike an optimal balance, we employ Low-Rank Adaptation (LoRA) \cite{Hu2022Lora}. We freeze the pre-trained weights $W_{0}\in\mathbb{R}^{d\times k}$ and inject trainable rank-decomposition matrices $B\in\mathbb{R}^{d\times r}$ and $A\in\mathbb{R}^{r\times k}$ specifically into the Query ($W_q$), Key ($W_k$), and Value ($W_v$) projection layers of each Transformer block. For an input vector $x \in \mathbb{R}^{k}$, the adapted forward pass for the output $h \in \mathbb{R}^{d}$ is formulated as:

\begin{equation}
h = W_0 x + \Delta W x = W_0 x + B A x ,
\end{equation}
where $W_0 \in \{W_q, W_k, W_v\}$ , $\Delta W = BA$ represents the incremental update with $A$ being initialized using a Gaussian distribution and $B$ initialized to zero.The variables $d, k$, and $r$ denote the output, input, and predefined rank dimensions, respectively, with $r \ll \min(d, k)$. This strategy reduces the trainable parameters for backbone to merely 2.36M (approx. 0.77\%), significantly lowering training costs while preserving the robust generalization of the LVM. Regarding deployment, while the ViT-Large backbone targets high-precision scenarios (e.g., ground stations), our framework maintains high scalability. As demonstrated in our experiments, the backbone can be seamlessly substituted with lighter variants for resource-constrained onboard platforms without altering the core architecture.

\subsection{Semantic Gated Residual Fusion (SGRF)}
The hierarchical architecture of ViT intrinsically entails that shallow and deep stages attend to distinct information. Although previous works in CVGL have demonstrated the effectiveness of two-layer fusion \cite{Tang2025R2PLoc}, recent findings suggest that multi-layer aggregation offers superior representation. We argue that existing two-layer fusion strategies often fail to capture fine-grained low-level textural cues. Furthermore, naive addition or concatenation risks introducing task-irrelevant noise from shallow layers into the final representation. Empirically, as shown in Table V, we observe that mid-level feature (Layer 20) outperform the final output. Thus, we establish Layer 20 as the primary anchor. Additionally, we utilize the final layer (Layer 24) for its rich semantic information, and Layer 16 to retain essential spatial details. Driven by these insights, we propose SGRF which adopts a ``guidance-and-gating'' strategy: it leverages high-level semantics to generate calibration weights that refine mid-level feature, and subsequently uses these refined feature to produce gating signals that filter out low-level noise.

Structurally, SGRF performs two sequential steps: semantic calibration and detail gating. Let $F_{l}, F_{m}, F_{h} \in \mathbb{R}^{C \times H \times W}$ (where $C, H, W$ denote the number of channels, height, and width, respectively) denote the low-, mid-, and high-level feature maps. Starting with the anchor feature, $F_{m}$ is transformed into the refined feature $F_m^{ref}$ via a residual block to enrich mid-level semantics. Specifically, $F_{m}$ passes through a $3\times3$ convolution ($\text{Conv}_{3\times3}$) and is added back to the original input:

\begin{equation}
    F_{m}^{ref} = F_{m} + \mathcal{B}(\text{Conv}_{3\times3}(F_{m})) ,
\end{equation}
where $\mathcal{B}$ denotes Batch Normalization and ReLU activation. Simultaneously, inspired by the Squeeze-and-Excitation (SE) mechanism \cite{Hu2018Squeeze}, the high-level feature $F_{h}$ undergoes Global Average Pooling (GAP) to squeeze spatial information. This is followed by a Multi-Layer Perceptron (MLP) to generate channel-wise calibration weights $W_{se}$: 
\begin{equation}
    W_{se} = \sigma(\mathbf{W}_{2}\delta(\mathbf{W}_{1}\text{GAP}(F_{h}))) .
\end{equation}
Here, $\sigma$ is the Sigmoid function, $\delta$ is ReLU, and $\mathbf{W}_{1}, \mathbf{W}_{2}$ are the weights of the MLP. Unlike standard SE blocks that perform self-recalibration, these weights are then applied to the refined mid-level feature via element-wise multiplication ($\odot$) to emphasize semantically relevant components. A $1\times1$ convolution ($\text{Conv}_{1\times1}$) then fuses the information to obtain the semantic-enhanced feature $F_{mh}$:
\begin{equation}
    F_{mh} = \text{Conv}_{1\times1}(F_{m}^{ref} \odot W_{se}) .
\end{equation}

The Detail Gating module incorporates a gating mechanism  to address the noise inherent in shallow layers, which often stems from insufficient semantic filtering during initial feature extraction. To prevent these task-irrelevant local textures from contaminating the final representation, we generate a spatial gate map guided by the semantically robust $F_{mh}$ instead of directly aggregating low-level features. Specifically, a $1\times1$ convolution followed by a Sigmoid function is applied to $F_{mh}$ to produce the gate map $M_{gate} \in \mathbb{R}^{C \times H \times W}$. This map adaptively determines the importance of each spatial position in the low-level feature $F_{l}$:
\begin{equation}
    M_{gate} = \sigma(\text{Conv}_{1\times1}(F_{mh})) ,
\end{equation}
\begin{equation}
    F_{out} = F_{mh} + (F_{l} \odot M_{gate}) .
\end{equation}
Through this mechanism, $F_{l}$ is injected into the final representation only when the semantic context deems it informative, effectively filtering out background noise. Designed for efficiency, SGRF introduces only 4.53 GFLOPs when processing 1024-dimension feature maps, ensuring negligible overhead.

\subsection{Sequential Aggregation (SA) Head}
To efficiently aggregate the fused 2D feature map into a compact global descriptor, we propose the SA Head. While Transformer-based heads offer global modeling capabilities, their quadratic complexity $O(L^2)$ incurs excessive computational overhead. Conversely, traditional CNN pooling methods struggle to capture long-range dependencies between distant landmarks.

The SA module adopts Mamba \cite{Gu2024Mamba}, a state-space model that achieves global receptive fields with linear complexity $O(L)$. Specifically, the input feature map $F_{fused} \in \mathbb{R}^{C \times H \times W}$ is flattened into a visual sequence $S_{in} \in \mathbb{R}^{L \times C}$ (where $L=HW$), following a row-major scan order. This sequence is processed by stacked Mamba blocks, which utilize the selective scan mechanism to efficiently model spatial dependencies across the flattened sequence.

Unlike standard NLP tasks that rely on the final token for prediction, visual cues in geo-localization are spatially distributed. Therefore, instead of extracting the last hidden state, we employ GeM Pooling \cite{Radenovic2018Fine} on the entire output sequence to generate the final representation,which is subsequently projected and normalized to the target embedding dimension. This design allows the model to adaptively focus on salient regions across the global context while maintaining high inference efficiency.


\begin{table*}[t]
\caption{Comparison of performance and efficiency on the University-1652 dataset. The best results are highlighted in \textbf{bold}. $\dagger$ indicates trainable parameters due to LoRA tuning.}
\label{tab:sota_university1652_efficiency}
\centering
\scriptsize
\setlength{\tabcolsep}{3pt}
\renewcommand{\arraystretch}{0.92}
\resizebox{\textwidth}{!}{
\begin{tabular}{llccccccc}
\toprule
\multirow{2}{*}{\textbf{Method}} 
& \multirow{2}{*}{\textbf{Backbone}} 
& \textbf{Params} 
& \textbf{FLOPs} 
& \multicolumn{2}{c}{\textbf{Drone $\to$ Satellite}} 
& \multicolumn{2}{c}{\textbf{Satellite $\to$ Drone}} 
& \multirow{2}{*}{\textbf{Avg.}~$\uparrow$} \\
\cmidrule(lr){5-6} \cmidrule(lr){7-8}
& & \textbf{(M)}~$\downarrow$ 
& \textbf{(G)}~$\downarrow$ 
& \textbf{R@1}~$\uparrow$ 
& \textbf{AP}~$\uparrow$ 
& \textbf{R@1}~$\uparrow$ 
& \textbf{AP}~$\uparrow$ 
& \\
\midrule
LPN \cite{Wang2021Each} & ResNet-50 & 62.4 & 36.8 & 75.93 & 79.14 & 86.45 & 74.49 & 79.00 \\
SDPL \cite{Chen2024SDPL} & Swin-T & 42.6 & 69.7 & 90.16 & 91.64 & 93.58 & 89.45 & 91.21 \\
MCCG \cite{Shen2024MCCG} & ConvNeXt-B & 56.7 & 51.0 & 89.40 & 91.07 & 89.93 & 95.01 & 91.35 \\
Sample4Geo \cite{Deuser2023Sample4Geo} & ConvNeXt-B & 87.6 & 90.2 & 92.65 & 93.81 & 95.14 & 91.39 & 93.25 \\
CCR \cite{Du2024CCR} & ViT-B & 156.6 & 160.6 & 92.54 & 93.78 & 91.80 & 95.15 & 93.32 \\
MEAN \cite{Chen2025Multi} & ConvNeXt-T & \textbf{36.5} & \textbf{26.2} & 93.55 & 94.53 & 96.01 & 92.08 & 94.04 \\
SHAA \cite{Chen2025SHAA} & SwinV2-B & 90.6 & 68.8 & 93.69 & 94.68 & 96.15 & 93.49 & 94.50 \\
DAC \cite{Xia2024Enhancing} & ResNet-50 & 96.5 & 90.2 & 94.67 & 95.50 & 93.79 & \textbf{96.43} & 95.10 \\
\midrule
\textbf{DINO-GFSA (Ours)} & \textbf{DINOv3-ViT-L} & 45.73$^\dagger$ & 134.3 & \textbf{95.68} & \textbf{96.34} & \textbf{96.29} & 95.56 & \textbf{95.97} \\
\bottomrule
\end{tabular}
}
\vspace{-0.2cm}
\end{table*}

\subsection{Loss Function}
Unlike prior methods that rely on a hybrid objective combining ID-level classification (e.g., Cross-Entropy) and metric constraints (e.g., Triplet Loss), we adopt a unified contrastive learning paradigm optimized solely via the InfoNCE loss \cite{Oord2018Representation}. Instead of learning fixed class boundaries for specific locations, this approach focuses on instance-level discrimination. It effectively maximizes the mutual information between matched pairs by pulling them closer in the embedding space while pushing away all other negative samples in the mini-batch. The loss function is defined as:
\begin{equation}
\mathcal{L}_{\text{InfoNCE}} = - \log \frac{\exp(\text{sim}(q, k_+) / \tau)}{\sum_{i=0}^{K} \exp(\text{sim}(q, k_i) / \tau)} ,
\end{equation}
where $q$ and $k_+$ denote the embeddings of the query and its corresponding positive sample, and $\{k_i\}_{i=0}^K$ represents the set of all keys in the mini-batch (including one positive and $K$ negatives). $\tau$ is the temperature hyperparameter, and $\text{sim}(\cdot)$ computes the cosine similarity.

\section{Experiments}

\subsection{Datasets and Evaluation Metrics}
To verify the versatility of our model across different operational requirements—specifically target localization and UAV self-positioning—we conduct evaluations on two distinct benchmarks:
\begin{itemize}
    \item \textbf{University-1652\cite{Zheng2020University}}: The first large-scale benchmark for multi-view UAV-satellite matching and region-level target localization. It contains approximately 50k images of 1,652 university buildings. We evaluate performance using Recall@K (R@K) and Average Precision (AP).
    \item \textbf{DenseUAV\cite{Dai2023Vision}}: Focused on self-positioning in GNSS-denied urban areas, this dataset consists of paired UAV and satellite images collected at 20m intervals across 14 universities. Besides R@K, we report SDM@1 (Spatial Distance Metric) , which incorporates spatial Euclidean distance to measure positioning accuracy beyond simple retrieval recall.
\end{itemize}

\begin{table}[t]
\caption{Comparison with SOTA methods on DenseUAV.}
\centering
\label{tab:sota_denseuav}
\footnotesize
\setlength{\tabcolsep}{2.5pt}
\resizebox{\columnwidth}{!}{
\begin{tabular}{lcccc}
\toprule
\textbf{Method} & \textbf{Backbone} & \textbf{R@1} & \textbf{R@5} & \textbf{SDM@1} \\
\midrule
DenseUAV Baseline \cite{Dai2023Vision} & ViT-S & 83.01 & 95.58 & 86.50 \\
MSBA \cite{Zhuang2021Faster} & ResNet50 & 46.13 & 64.22 & 52.64 \\
LPN \cite{Wang2021Each} & ViT-S & 71.77 & 90.13 & 77.95 \\
FRSA \cite{Dai2022Transformer} & ViT-S & 81.21 & 94.55 & 85.11 \\
DINOv2 based method \cite{Yang2025Dinov2} & DINOv2-B & 86.27 & 96.83 & 88.87 \\
MCCG \cite{Shen2024MCCG} & ConvNext-T & 89.19 & 96.87 & 90.99 \\
SHAA \cite{Chen2025SHAA} & SwinV2-B & 93.69 & 98.76 & 94.91 \\
\midrule
\textbf{DINO-GFSA(Ours)} & DINOv3-ViT-L & \textbf{97.17} & \textbf{99.57} & \textbf{97.68} \\
\bottomrule
\end{tabular}
}
\end{table}

\subsection{Implementation Details}
We implement the model using PyTorch on an NVIDIA RTX 6000 GPU. The input images are set to $224 \times 224$. We fine-tune the backbone using LoRA ($r=16, \alpha=32$) and train for 80 epochs with a batch size of 128. Optimization is performed using AdamW (weight decay $5 \times 10^{-4}$) and a Cosine Annealing scheduler. The learning rates are initialized at $2.8 \times 10^{-4}$ for the backbone and $5 \times 10^{-5}$ for other modules. The temperature $\tau$ for the InfoNCE loss is set to 0.07. Data augmentation techniques including random horizontal flipping, color jittering, and random erasing are applied during training.
\subsection{Comparison with State-of-the-Art Methods}

\textbf{Results on the University-1652 dataset}, as illustrated in Table I, demonstrate the superiority of our proposed method. For UAV-to-satellite retrieval, we achieve 95.68\% on R@1 and 96.34\% on AP, surpassing the previous SOTA by 1.01\% and 0.84\%, respectively. In the satellite-to-UAV task, our method attains 96.29\% on R@1 and 95.56\% on AP, consistently ranking among the top-tier methods. Notably, our average performance across these four metrics reaches 95.97\%, outperforming the second-best method DAC by 0.87\%.

In terms of efficiency, our architecture is based on a large-scale foundation model (ViT-Large). Consequently, our computational cost is 134.3 GFLOPs. While this is higher than mainstream methods like DAC (96.5 GFLOPs) or SHAA (90.6 GFLOPs) and only lower than CCR (160.6 GFLOPs), it is a necessary trade-off to explore the performance upper bound of LVMs in geo-localization. Importantly, thanks to the LoRA strategy, we strictly limit the trainable parameters to just 45.73M, which is the third lowest among all compared methods. This makes our model training-efficient despite the high inference capacity.

\textbf{Results on the DenseUAV benchmark}, as illustrated in Table II, provide compelling evidence of our method’s superiority. We attain 97.17\% on R@1, 99.57\% on R@5, and 97.68\% on SDM@1, surpassing the previous SOTA (SHAA) by 3.48\%, 0.81\%, and 2.77\%, respectively. Crucially, we emphasize that this performance leap is not solely reliant on the backbone scale. As detailed in the ablation study (Table VI), even our Base variant outperforms the previous SOTA, confirming that the improvements primarily stem from our architectural innovations. We attribute this to the SGRF module's ability to handle the specific characteristics of DenseUAV. Since dense sampling creates high inter-sample redundancy, the model requires fine-grained discrimination to avoid mismatching. The SGRF module’s deep fusion of spatial-semantic feature meets this demand perfectly, while the Large backbone further pushes the performance boundary to its limit.

\subsection{Ablation Study}
In this section, we present an extensive ablation study to validate the individual contributions of our proposed strategies. All experiments in this section are conducted on the DenseUAV dataset to standardize the evaluation criteria.

\begin{table}[b]
    \centering
    \caption{Ablation study of different components on the DenseUAV dataset. The values in parentheses indicate the improvement over the previous baseline.}
    \label{tab:ablation_components}
    \resizebox{\linewidth}{!}{
    \begin{tabular}{l|cc}
        \toprule
        \textbf{Method} & \textbf{Recall@1} & \textbf{SDM@1}  \\
        \midrule
        Baseline (backbone + GeM) & 94.24 & 95.01 \\
        \midrule
        + SA (1 Mamba layer) & 95.47 \small{(+1.23)} & 96.41 \small{(+1.30)} \\
        + SA (2 Mamba layers) & 96.35 \small{(+0.88)} & 97.02 \small{(+0.61)} \\
        \textbf{+ SA + SGRF (Proposed)} & \textbf{97.17} \small{(+0.82)} & \textbf{97.68} \small{(+0.66)} \\
        \bottomrule
    \end{tabular}
    }
\end{table}

\begin{table}[t]
\caption{Comparison of trainable parameters and performance (R@1, SDM@1) across different fine-tuning strategies.}
\label{tab:ablation_sdm}
\centering
\footnotesize
\setlength{\tabcolsep}{3pt}
\resizebox{\columnwidth}{!}{
\begin{tabular}{l|c|cc} 
\toprule
\textbf{Fine-tuning Strategy} & \textbf{Params (M)} & \textbf{R@1} & \textbf{SDM@1} \\
\midrule
Frozen / 0 layers  & 43.37  & 92.36 & 94.14 \\
Unfreeze 4 layers  & 93.85  & 92.41 & 94.18 \\
Unfreeze 8 layers  & 144.22 & 94.94 & 95.97 \\
Unfreeze 12 layers & 194.59 & 96.35 & 97.05 \\
Full Fine-tuning   & 346.50 & 96.40 & 97.12 \\
\midrule
\textbf{LoRA} & \textbf{45.73} & \textbf{97.17} & \textbf{97.68} \\
\bottomrule
\end{tabular}
}
\end{table}

\paragraph{\textbf{Ablation on components}}
As shown in Table III , we examine the contribution of each core module on the DenseUAV dataset. The baseline (DINOv3 with GeM pooling) achieves 94.24\% R@1 and 95.01\% SDM@1. Replacing GeM with a single-layer SA Head yields a notable gain, reaching 95.47\% R@1 (+1.23\%) and 96.41\% SDM@1 (+1.30\%). Stacking a second Mamba block further improves performance to 96.35\% R@1 and 97.02\% SDM@1 , indicating that deeper sequential modeling better captures complex spatial dependencies. Finally, the integration of SGRF refines the results to 97.17\% R@1 and 97.68\% SDM@1 , validating the effectiveness and synergy of the DINO-GFSA framework.

\paragraph{\textbf{Evaluation on DINOv3 backbone and LoRA}} 
Table~\ref{tab:ablation_sdm} compares LoRA against gradually unfreezing layers from the last block forward. While full fine-tuning improves R@1 by 4.04\% over the frozen baseline, it incurs an 8-fold increase in trainable parameters. In contrast, LoRA demonstrates superior efficiency: with only 2.36M additional parameters, it achieves a 4.81\% improvement, outperforming even the full fine-tuning strategy in both accuracy and efficiency.
\begin{table}[t]
\centering
\caption{Ablation study of different feature layers and fusion strategies on DINOv3 backbone.}
\label{tab:ablation_fusion}
\footnotesize
\setlength{\tabcolsep}{3pt}
\resizebox{\columnwidth}{!}{
\begin{tabular}{llcc}
\toprule
\textbf{Category} & \textbf{Settings} & \textbf{R@1} & \textbf{SDM@1} \\
\midrule
\multirow{3}{*}{Single Layer} 
  & Low Layer(16) & 95.24 & 96.18 \\
  & Mid Layer(20) & 96.35 & 97.02 \\
  & High Layer(24) & 95.54 & 96.46 \\
\midrule
\multirow{2}{*}{Partial Combination} 
  & Layer Low + Mid & 96.78 & 97.31 \\
  & Layer Mid + High & 96.78 & 97.33 \\
\midrule
\multirow{3}{*}{Fusion Strategy} 
  & Addition Fusion & 96.59 & 97.22 \\
  & Concat Fusion & 96.61 & 97.38 \\
  & \textbf{Proposed} & \textbf{97.17} & \textbf{97.68} \\
\bottomrule
\end{tabular}
}
\end{table}

\begin{table}[b]
  \centering
  \caption{Ablation study evaluating performance across different backbone sizes. Lightweight variants (Small/Base) utilize a single-layer SA Head for efficiency.}
  \label{tab:backbone_efficiency}
  \renewcommand{\arraystretch}{1.1} 
  \resizebox{\linewidth}{!}{
    \begin{tabular}{lcccc}
      \toprule
      \textbf{Backbone Size} & \textbf{Params (M)} & \textbf{FLOPs (G)} & \textbf{R@1 (\%)} & \textbf{SDM@1 (\%)} \\
      \midrule
      DINOv3-Small   & 28.3  &  9.98  & 89.06 & 91.23 \\
      DINOv3-Small+ & 33.3  &  12.86 & 91.68 & 93.25 \\
      \textbf{DINOv3-Base}   & \textbf{102.0}  & \textbf{39.4}  & \textbf{94.72} & \textbf{95.78} \\
      DINOv3-ViT-L           & 346.5           & 134.3           & 97.17 & 97.68 \\
      \bottomrule
    \end{tabular}%
  } 
  \vspace{-0.2cm}
  \\[2pt]
\end{table}

\paragraph{\textbf{Effectiveness of SGRF}} 
Table V validates SGRF's ability to bridge the semantic gap. First, the mid-level feature (Layer 20) serves as a strong anchor (96.35\% R@1). Second, applying our fusion to just two layers (partial combination) already improves accuracy to 96.78\%, notably surpassing standard strategies like addition (96.59\%) and concatenation (96.61\%). This suggests that without semantic guidance, naive integration is suboptimal even with more layers. Finally, the full three-layer SGRF further boosts performance to 97.17\%, establishing a clear lead of 0.56\% over Concatenation. This confirms that while dual-layer fusion is beneficial, the holistic, semantically calibrated integration of all three levels is essential for maximizing robustness.

\paragraph{\textbf{Scalability and Onboard Deployment}}
As detailed in Table VI, the trade-off between accuracy and computational cost can be flexibly managed.
The DINOv3-Vit-Large variant pushes the performance boundary, making it ideal for high-precision scenarios such as ground station processing.
Conversely, for resource-constrained onboard platforms, the DINOv3-Base variant demonstrates exceptional efficiency. With only 39.4 GFLOPs, it significantly undercuts the computational cost of the previous SOTA method SHAA (90.6 GFLOPs) while still outperforming it in accuracy (94.72\% vs 93.69\%). This confirms that our framework is highly scalable and can be tailored to specific deployment constraints.

\begin{figure}[t]
\centering
\includegraphics[width=\linewidth]{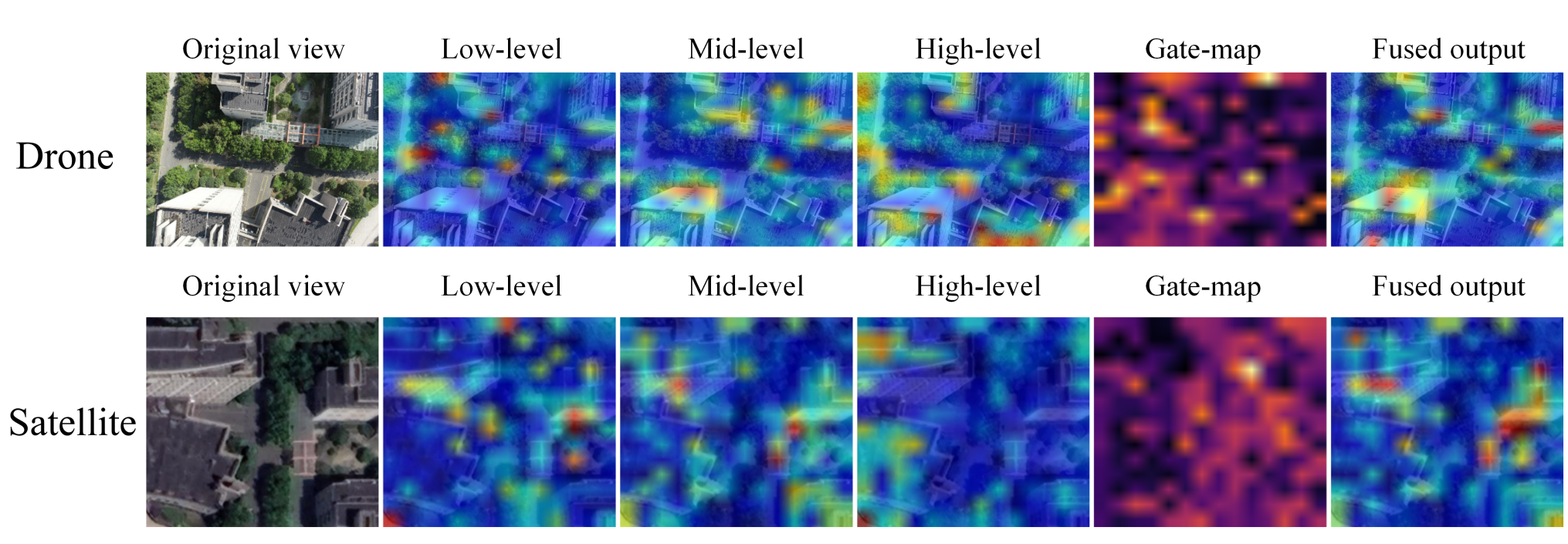} 
\caption{Visualization of feature evolution. Top: UAV view; Bottom: Satellite view. Columns from left to right: Original image, Low-level (noisy), Mid-level, High-level features, the gate map learned by SGRF, and the Final fused output. Note how the gate map (dark regions) effectively suppresses environmental clutter (e.g., trees) while highlighting discriminative building structures in the fused output.}
\label{fig:vis}
\vspace{-0.5cm}
\end{figure}

\subsection{Visualization}


Fig.~\ref{fig:vis} visualizes feature evolution through the SGRF module. In the UAV view (top), the learned gate map suppresses environmental clutter (e.g., vegetation) in low-level feature, focusing the fused output on salient building structures. Conversely, for the satellite view (bottom), SGRF compensates for the lack of spatial precision in high-level semantics by adaptively incorporating sharp geometric cues from shallower layers. These results confirm DINO-GFSA’s ability to filter noise while preserving discriminative landmarks for accurate matching.


\section{Conclusion}
In this paper, we propose DINO-GFSA, a robust framework for UAV CVGL. By synergizing the DINOv3 backbone with LoRA, we successfully harness the power of LVMs while keeping training costs low. Furthermore, the proposed SGRF module and SA Head effectively resolve the dilemmas of feature fusion and aggregation. Extensive experiments validate the superiority of our method across multiple benchmarks. Our framework demonstrates high scalability: the Large variant excels in precision for ground stations, while the Base and Small+ variants offer a viable solution for efficient onboard deployment. Future research focuses on real-world flight tests to optimize performance under dynamic, physical constraints.

\end{document}